\documentclass{Interspeech2024}

\interspeechcameraready

\title{A Unit-based System and Dataset \\for Expressive Direct Speech-to-Speech Translation}

\name[affiliation={1}]{Anna}{Min*}
\name[affiliation={1}]{Chenxu}{Hu*}
\name[affiliation={2}]{Yi}{Ren}
\name[affiliation={1}]{Hang}{Zhao}

\address{
  $^1$Tsinghua University, China\\
  $^2$ByteDance, China}
\email{man20@mails.tsinghua.edu.cn, hu-cx21@mails.tsinghua.edu.cn, ren.yi@bytedance.com, hangzhao@tsinghua.edu.cn}


\begin{document}

\maketitle

\begin{abstract}
{\begin{NoHyper}\let\thefootnote\relax\footnotetext{$^{*}$ Equal contribution.}\end{NoHyper}}
Current research in speech-to-speech translation (S2ST) primarily concentrates on translation accuracy and speech naturalness, often overlooking key elements like paralinguistic information, which is essential for conveying emotions and attitudes in communication. To address this, our research introduces a novel, carefully curated multilingual dataset from various movie audio tracks. Each dataset pair is precisely matched for paralinguistic information and duration. We enhance this by integrating multiple prosody transfer techniques, aiming for translations that are accurate, natural-sounding, and rich in paralinguistic details. Our experimental results confirm that our model retains more paralinguistic information from the source speech while maintaining high standards of translation accuracy and naturalness.
\end{abstract}

\section{Introduction}

Speech-to-speech translation (S2ST) enables the translation of spoken language into another spoken language, significantly enhancing communication between different language speakers. Traditional S2ST systems~\cite{lavie1997janus,nakamura2006atr,huang2023holistic} rely on a pipeline of automatic speech recognition (ASR), machine translation (MT), or speech-to-text translation (S2T), followed by text-to-speech synthesis (TTS). While speech translation traditionally involves converting speech to text or vice versa in different languages, recent developments have shifted towards an end-to-end S2ST system~\cite{lee-etal-2022-direct,popuri2022enhanced,jia2022translatotron,jia2019direct,lee2021textless}. These systems minimize error propagation between ASR and MT, resulting in a streamlined process with reduced computational costs, particularly advantageous for languages without a written form.

In the realm of expressive speech-to-speech translation (S2ST)~\cite{lee2022direct,popuri2022enhanced,huang2023transpeech,huang2023holistic,barrault2023seamless}, some research focuses on intonation transfer, utilizing statistical word alignment to transfer source intonation characteristics to the target language. Many methods~\cite{do2016preserving,do2018sequence,waibel2023face} evolved to include word emphasis transfer, ultimately leading to sequence-to-sequence models for simultaneous emphasis and content translation. Despite the progress, these approaches only focused on individual expression elements and did not fully capture the emotional aspects of speech.


However, a notable challenge in direct S2ST with style transfer is the scarcity of paired data where the source and target speech have the same speaker. Some works~\cite{diwan2023unit,nachmani2023translatotron} use non-parallel data, but they lack ground truth for human evaluation, limiting further study. To address this, we introduce a novel, carefully curated multilingual dataset from diverse movie audio tracks. This dataset, primarily consisting of paired Spanish-English data from clear, emotionally rich dialogues in movies and TV shows, captures nuanced emotional variations often missed in standard speech synthesis data.


To extract fine-grained emotional information, we propose a model that learns global style and local pitch features, as discrete speech representations may lose some paralinguistic information. Thus, our direct S2ST method can preserve more paralinguistic characteristics. In addition, our direct S2ST model translates between languages without intermediate text, using discrete units instead. 

Our contributions are as follows:
\begin{itemize}
    \item We introduce the first dataset for training paired speech emotion translation from movies and TV shows with multiple audio tracks and provide a scalable automatic pipeline to expand this dataset for future research.
    \item We propose a novel approach to direct S2ST with style transfer, integrating global style and local pitch transfer. This method preserves intricate emotional characteristics without needing text as an intermediary while maintaining translation accuracy.
    \item Our experiments demonstrate that our method transfers emotions while maintaining comparable translation quality.
\end{itemize}

\vspace{-0.5cm}
\begin{figure}[ht]
    \centering
    \includegraphics[width=\linewidth]{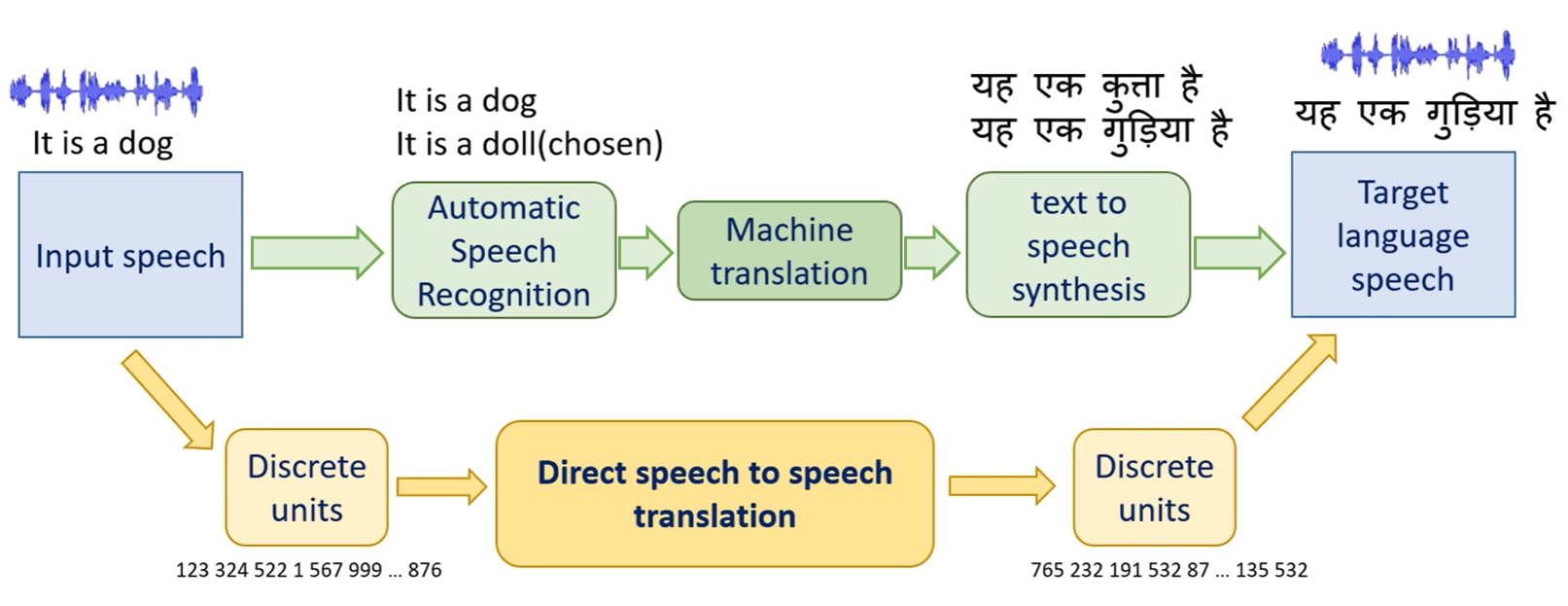}
    \caption{Direct speech-to-speech translation system compared with cascaded speech-to-speech translation system: The green-colored pipeline above represents the traditional cascaded approach, which requires text as an intermediary. The approach below involves a discrete unit translation method, eliminating the need for text as an intermediary.}
    \label{fig:my_label}
\end{figure}

\vspace{-0.6cm}

\section{Movie Dataset}

In this section, we detail the construction and processing of the movie dataset, which is crucial for advancing research in speech-to-speech translation between English and Spanish and sets the stage for future research in producing emotionally paired multilingual speech datasets. To leverage additional multilingual resources available on the internet in the future, we have developed an automated pipeline, unlike ~\cite{huang2023holistic}, which requires manual annotations to obtain high-quality paired training data. 

\subsection{Dataset Source}
Our dataset, a substantial collection of approximately 300 hours of paired English-Spanish television series and movie audio, is carefully curated to facilitate advanced translation model development. 
Our dataset comprises content from the following sources: ``Money Heist'' seasons 1-5, ``Elite'' seasons 1-4, 59 Disney movies, ``Dragon Ball Z'', 24 ``James Bond'' Collection movies, 10 superhero series movies, ``Shrek'' movies 1-4, ``Harry Potter'' movies 1-8, and ``Poltergeist'' movies 1-3. The videos feature both English and Spanish audio tracks, with actors ensuring emotion and timing alignment through diligent dubbing. In this work, we focus on speech translation.
Our dataset maintains a certain degree of emotional consistency between English and Spanish dubbing counterparts. This uniformity ensures standardization, which is critical for voice recognition and translation models.

\subsection{Dataset Construction}
We start constructing our dataset by converting subtitle SRT files into structured CSV files. This conversion process was enhanced with additional rules based on the SRT format (such as speaker information and sentence pause indicators) to eliminate irrelevant or inconsistent data. Additionally, we merge consecutive sentences from the same speaker using~\cite{desplanques2020ecapa} to ensure speaker consistency, consolidating them into a single data point. This improves dataset coherence and contextual relevance, crucial for training models on realistic dialogue patterns and maintaining narrative continuity.

\subsubsection{Filter by ASR Accuracy and Duration}
Next, we subject all audio files to a denoising process using a noise suppression library\footnote{https://github.com/xiph/rnnoise} based on a recurrent neural network, significantly improving audio clarity. This clarity is vital for the subsequent step of automatic speech recognition (ASR) using Azure\footnote{https://azure.microsoft.com/en-us/products/ai-services/ai-speech}, as cleaner audio leads to more accurate transcription. We select segments where the ASR output is closely aligned with the subtitles, choosing the top 80\% of segments with a word error rate more significant than 0.6 to ensure dataset accuracy. Additionally, we carefully filter segments based on appropriate sentence lengths, optimizing the dataset for practical training while maintaining contextual richness. Specifically, we exclude segments shorter than 3 seconds or longer than 15 seconds. All audio is processed at the sampling rate of 16000.


\begin{figure}[ht]
    \centering
    \includegraphics[width=0.8\linewidth]{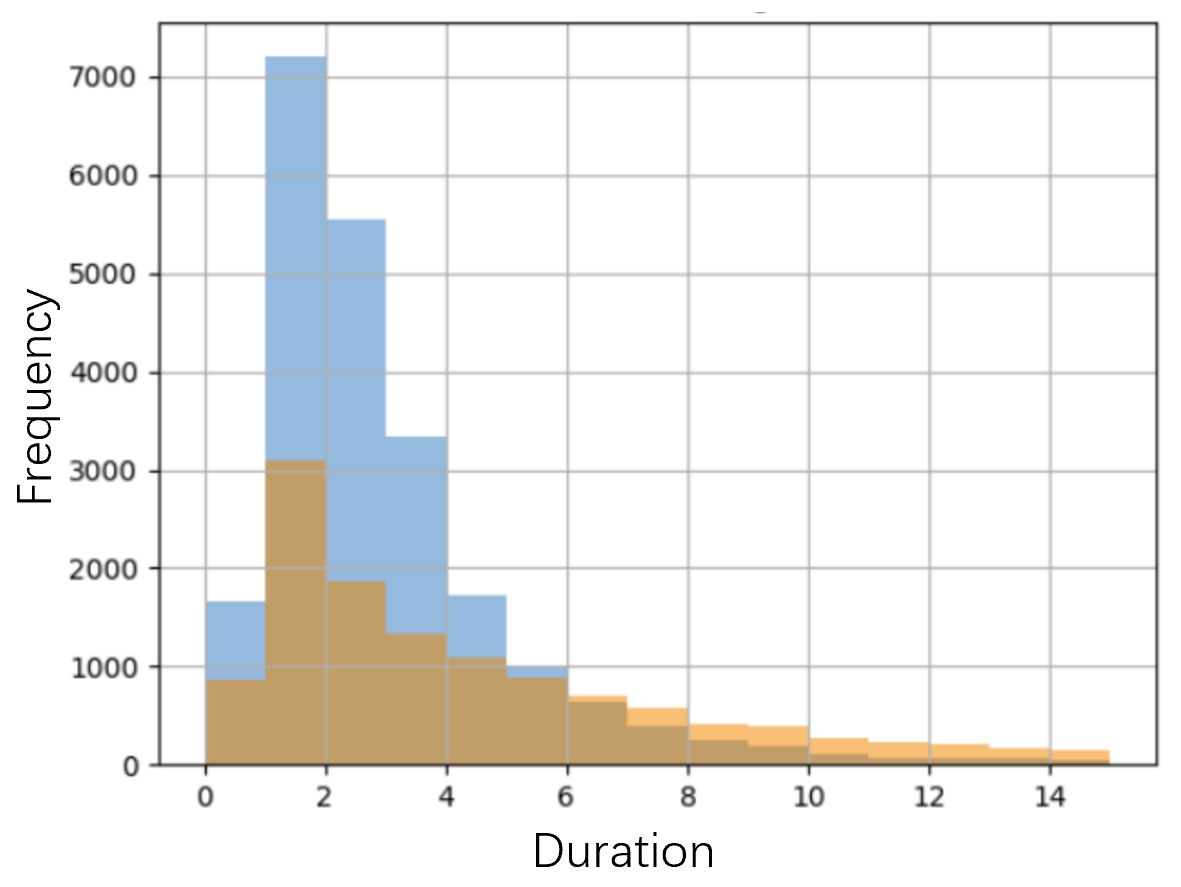}
    \caption{This is the length Distribution of the utterances, and the yellow ones denote the utterances that have a word error rate under 40\%. There are 12610 utterances in total. The maximum duration is 244.250s, while the minimum duration is 0.833s. The average duration of utterances is 5.096s.}
    \label{fig:my_label}
\end{figure}


\subsubsection{Filter by Speakers of the Utterances}
Additionally, we analyze the English audio segments. This involved matching English segments with corresponding Spanish segments within the same TV show, ensuring consistency in contextual and emotional content. We prioritize segments where the cosine similarity between the matched English and Spanish audio was below 0.5, ensuring diversity in the dataset that challenges and thus strengthens the robustness of the translation models. Furthermore, we impose a criterion that only lists at least five remaining segments that would be saved, ensuring a meaningful sample size for model training.


\section{Method}
To extract fine-grained emotional information from these paired datasets, we develop a model to transfer the emotion of the reference speech. Our S2ST system comprises three components. Initially, speech from one language is converted into discrete units for direct speech-to-speech translation. Subsequently, speaker identification is extracted from the speech. Illustrated in Figure 3, we introduce a unit-hifigan-based emotion transfer model, where speech in the target language, enriched with the appropriate emotions, is synthesized.

\subsection{Obtain Discrete Units}
In the first stage, we extract discrete units using a process inspired by the HuBERT~\cite{hsu2021HuBERT} framework as in ~\cite{popuri2022enhanced}, which employs self-supervised learning techniques for speech representation. HuBERT leverages K-means clustering on its intermediate representations or Mel-frequency cepstral coefficient (MFCC) features~\cite{davis1980comparison} in the initial iteration to categorize masked audio segments into discrete labels. By pre-training a HuBERT model on an unlabelled speech corpus in the target language, we can encode target speech into continuous representations for every 20ms frame. Subsequently, a K-means algorithm is applied to these representations to generate K cluster centroids. These centroids are instrumental in encoding target utterances into sequences of cluster indices at the same 20ms interval.

We follow \cite{popuri2022enhanced} to encode the target speech into a vocabulary of 1000 discrete units. The models for HuBERT and K-means are derived from a combination of unlabeled English, Spanish, and French speech data sourced from the VoxPopuli~\cite{wang2021voxpopuli} corpus. Our focus is solely on encoding English and Spanish target speech.

The second stage involves processing the discrete units obtained from the first stage. Due to the high length of the original unit, we follow~\cite{popuri2022enhanced} to adopt a strategy to condense continuous repetitions of the same unit into a single unit. This approach not only streamlines the dataset but also aids in reducing computational complexity. In the third and final stage, these condensed units are expanded back to their original form during the Unit-to-Waveform conversion process.

\subsection{Get Speaker ID}


In our approach, we extract speaker embeddings from each utterance to capture unique vocal characteristics. We compute cosine similarities between these embeddings to assign each utterance's speaker as a pseudo label.

We organize these embeddings and cosine similarities into dedicated files. This systematic organization improves data accessibility and facilitates comparative analysis across languages, ensuring easy retrieval and analysis of vocal features for further research.

Utilizing this approach, we can identify and extract speaker IDs from the dataset effectively. The speaker ID provides a unique identifier for each speaker, enabling us to track and analyze individual vocal characteristics across different linguistic contexts. This is particularly valuable in speech-to-speech translation research, where understanding and preserving individual speaker characteristics is essential for generating accurate and natural translations.


\begin{figure}[ht]
    \centering
    \includegraphics[width=\linewidth]{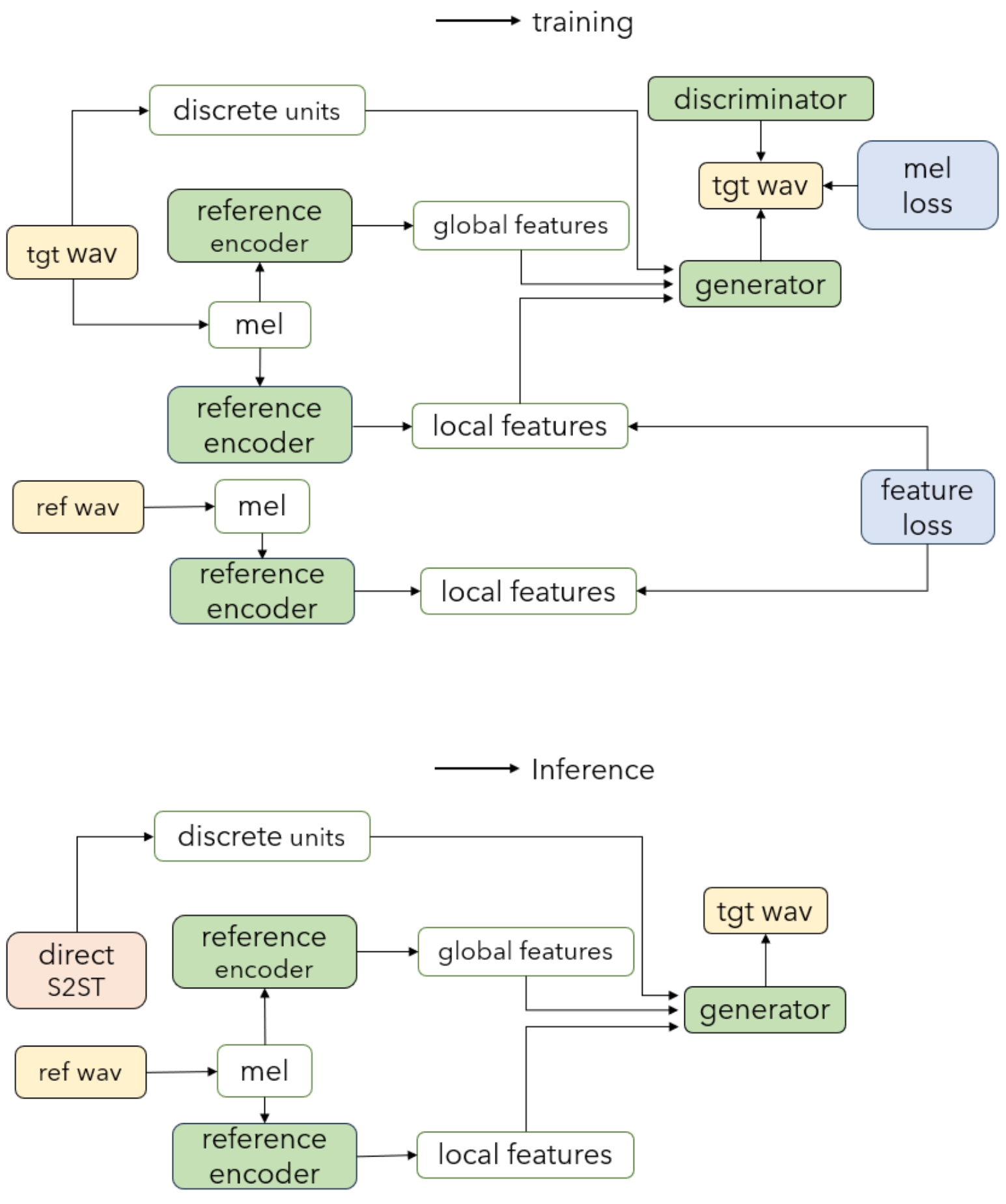}
    \caption{Our unit-HiFi-GAN-based voice style transfer model extracts local features from audio during both the training and inference stages. During inference, translated units and their corresponding reference waveforms are inputted into the model to synthesize the corresponding audio.}
    \label{fig:my_label}
\end{figure}


\subsection{Unit2Wav Synthesis}
In the third part of our method, we focus on synthesizing speech in a different language from the voice and translating discrete representations. This presents two primary challenges: firstly, maintaining high audio quality after dataset denoising, and secondly, addressing the tonal differences between matched audio in different languages, which create a gap that complicates the direct computation of Mel-spectral loss.

To tackle these issues, as shown in Figure 3, we employ a unit-based variant of HiFi-GAN~\cite{kong2020hifigan}, termed ``unit-HiFiGAN''. The structure is inspired mainly by HiFi-GAN, which excels in handling signals of varying periodicity in speech by employing multiple smaller sub-discriminators. These sub-discriminators individually process different periodic patterns, resulting in superior performance. Additionally, this model architecture allows for parallel processing of these patterns, enhancing computational efficiency. Controllable Text-to-Speech (TTS) has developed in two main directions: global and fine-grained style transfer~\cite{wang2018style,skerryryan2018endtoend,ren2022fastspeech,sun2020fullyhierarchical}. Global style transfer, encapsulating overall speech attributes into a single embedding, contrasts with fine-grained style transfer, which captures local prosodic features but faces alignment challenges. Global style transfer is more adaptable to non-parallel scenarios, so we leverage it in our S2ST framework following~\cite{skerryryan2018endtoend}.

Our innovation primarily lies in two areas. First, addressing the challenge of preserving high audio quality post-denoising, our model can be trained on high-quality monolingual datasets. This initial stage establishes a foundation for quality. Subsequently, we train the model and discriminator predictors on a mixed dataset, further refining the system.

Regarding the second challenge of tonal differences between matched audio in various languages, our approach includes predicting the speaker (spk) attributes from the outputs processed by the reference (ref) encoder. We then calculate a loss function based on this prediction, aiming to minimize the non-timbral features in the ref encoder's output. Acknowledging the typically deeper tones in Spanish speech, we input and output both Spanish-to-English and English-to-Spanish translations to mitigate potential model biases. Additionally, we integrate a pitch predictor and an unvoiced/voiced predictor into our system.

These strategic innovations in our method not only address the inherent challenges in cross-lingual speech synthesis but also push the boundaries of what's achievable in terms of audio quality and linguistic versatility.

\section{Experiments}

\subsection{Experimental Setup}

\subsubsection{Dataset Setup}
In addition to the movie dataset we introduced, to enhance the audio quality, we also utilize the supplementary training material from~\cite{wang2021voxpopuli,Kahn_2020,veaux2017english,ljspeech17}, which comprises over 400K hours of high-quality audio. As they lack paired translated audio, we employ the same audio during training as a substitute for translated audio. We use English as the target language and Spanish as the reference language.

\subsubsection{Human Evaluation Protocol}
To measure expressivity preservation, we adopt the protocols~\cite{huang2023holistic} proposed, focusing on four specific aspects of expressiveness. This system, based on established methodologies in the field, categorizes expressiveness into four core aspects: emphasis, intonation, rhythm, and emotion. These categories were selected based on internal qualitative research, which pinpointed them as critical for preserving expressiveness in speech translation.

Among these aspects, emphasis, intonation, and rhythm are related to more localized or prosodic features of speech. These elements play a crucial role in conveying the subtleties of spoken language. On the other hand, emotion is treated as the most ``global'' aspect, encompassing the overall feeling or mood conveyed by the speech. It's important to note that while we assess naturalness in our translations, it is conducted in a separate study and thus not included as a core expressiveness aspect in this experimental setup. 

To evaluate emotion expression and objectively rate the performance of our model across the identified expressivity aspects, we recruit 10 participants to assess multiple pairs of results containing our method and baseline outcomes through questionnaires following ~\cite{huang2023holistic}. Each pair is evaluated using four different scores, and the average rating across all 10 participants is calculated. Their evaluations are crucial in providing an unbiased assessment of our model's capability to preserve expressivity in speech-to-speech translation.

\subsubsection{Model Setup and Training Details}
In our expressive S2ST system, we use the open-source S2T model in the Fairseq toolkit~\cite{ott2019fairseq,wang2020fairseq,wang2021fairseq}. We use the pretrained Es-En model they provide. 

Regarding the unit-to-wav model, our configuration includes several key training parameters. We initiate training from scratch with a learning rate of $2e^{-4}$, a learning rate decay of $0.999$, and an inverse square root learning rate scheduler with the warmup. We employ the Adam optimizer with beta values of (0.8, 0.99) and set a dropout probability of 0.1 in both the reference encoder and prosody encoder. Regarding the training process, we supplement the dataset with additional speaker IDs corresponding to the respective datasets and directly mix them with our dataset for training, giving each utterance the same weight. During evaluation, we assess performance solely on our dataset.

For our baseline model, we utilize a vanilla implementation of the HiFi-GAN in ~\cite{polyak2021speech}, trained on the LJ Speech dataset~\cite{ljspeech17}. This choice of baseline ensures a dependable comparison to evaluate our model's performance in emotional voice conversion, leveraging HiFi-GAN's established capability in producing high-quality speech audio.

\subsection{Main Results}
\subsubsection{Preserve Emotion}
Our experimental results provide compelling evidence of the superiority of our model over traditional vanilla unit-based TTS systems, particularly in the realms of emphasis, intonation, and rhythm. These aspects are critical in achieving natural-sounding, expressive speech synthesis, a goal that has remained elusive in many existing TTS technologies.

\begin{table}[ht]
\caption{We assess the Cascade system's performance across various aspects of speech. Specifically, we evaluate the degree of emotional voice conversion using the criteria proposed by~\cite{huang2023holistic}.}
    \centering
    \resizebox{\columnwidth}{!}{%
    \begin{tabular}{lcccc}
    \toprule
    \textbf{System} & \textbf{Emotion$\uparrow$} & \textbf{Emphasis$\uparrow$} & \textbf{Intonation$\uparrow$} & \textbf{Rhythm$\uparrow$} \\
    \midrule
    Vanilla Unit-TTS & 2.03 & 2.68 & 2.46 & 2.30 \\
    Holistic Cascade & 3.58 & 3.26 & 3.17 & 3.56  \\
    \bottomrule
    \end{tabular}%
    }
    
    \label{tab:holistic_cascade}
\end{table}

\textbf{Emotion and Emphasis:} The performance of our model in replicating the emotion and the emphasis in the speech was markedly superior. This was quantitatively measured using a set of metrics designed to capture the degree of emphasis correctly copied from the source material. Our model showed an improvement of 21\% over traditional vanilla unit-TTS, indicating a more dynamic and contextually accurate speech synthesis. Regarding the baseline, due to the lack of additional linguistic information in discrete units, the speech tones are nearly uniform.

\textbf{Intonation:} Intonation, a vital component in conveying emotions and questions in speech, was another area where our model excelled. Using a specialized intonation accuracy index, we observe that our model's ability to mimic the natural intonation patterns of human speech surpassed that of vanilla unit-TTS by 29\%. This improvement is indicative of the model's sophisticated understanding of speech patterns and its ability to generate more human-like, expressive speech.

\textbf{Rhythm:} In terms of replicating the natural rhythm of speech, our model again outperformed the vanilla unit-based TTS. The rhythm conformity score, which measures how closely the synthesized speech matches the rhythm of natural speech, was 55\% higher in our model. This result underscores our model's advanced capability in capturing and reproducing the subtle temporal characteristics of speech, which are essential for naturalness and expressiveness.

These results were further corroborated by subjective evaluations, in which a panel of listeners rated our model's outputs as significantly more natural and expressive than those generated by the vanilla unit-TTS system.

\subsubsection{Translation Quality}

\begin{table}[ht]
\caption{The BLEU of different system's output compared to the English groundtruth. ``S2ST'' denotes using the translation result as the input of Unit-TTS/Unit2Wav.}
    \centering
    \begin{tabular}{lcc}
    \toprule
    \textbf{System} & \textbf{S2ST$\uparrow$} & \textbf{GT input$\uparrow$}  \\
    \midrule
    Vanilla Unit-TTS & 29.2 & 81.0  \\
    Ours & 28.3 & 74.6 \\
    GT & - & 78.2 \\
    \bottomrule
    \end{tabular}%
    
    \label{tab:holistic_cascade}
\end{table}

To measure the ability of our model to keep the high audio quality while maintaining both emotional expressions, we conduct Automatic Speech Recognition by Azure\footnote{https://azure.microsoft.com/en-us/products/ai-services/ai-speech} on the translations generated by our model and compare them to the ground truth results. The BLEU score achieved 74.6, whereas the ground truth yielded a BLEU score of 78.2. These results demonstrate the superior performance of our approach in terms of both fluency and accuracy in preserving the original content. The baseline vanilla unit-TTS exhibits a higher BLEU score, attributed to the fact that the LJ Speech training set comprises passages read by a single speaker. In contrast, our movie dataset consists of numerous noisy clips. This disparity is reasonable and highlights a limitation of our method. Future research endeavors could work on this challenge.

\section{Conclusion}
Direct S2ST faces challenges due to data scarcity. To tackle this, we introduce the first training dataset for expressive speech translation and propose a model for emotion translation, advancing S2ST. Our approach integrates pitch and global style transfer, enabling real-time improvements without relying on text intermediaries. This innovative method ensures high-quality translations while maintaining stylistic fidelity to the source speech. The dataset includes paired Spanish-English data from movie and TV show dialogues, capturing nuanced emotional variations often overlooked in conventional speech synthesis datasets. These advancements extend the capabilities of direct S2ST, paving the way for future developments in expressive speech synthesis and style transfer. Future work may expand our pipeline to include additional language pairs.

\bibliographystyle{IEEEtran}
\bibliography{template}

\begin{thebibliography}{10}
\providecommand{\url}[1]{#1}
\csname url@samestyle\endcsname
\providecommand{\newblock}{\relax}
\providecommand{\bibinfo}[2]{#2}
\providecommand{\BIBentrySTDinterwordspacing}{\spaceskip=0pt\relax}
\providecommand{\BIBentryALTinterwordstretchfactor}{4}
\providecommand{\BIBentryALTinterwordspacing}{\spaceskip=\fontdimen2\font plus
\BIBentryALTinterwordstretchfactor\fontdimen3\font minus \fontdimen4\font\relax}
\providecommand{\BIBforeignlanguage}[2]{{%
\expandafter\ifx\csname l@#1\endcsname\relax
\typeout{** WARNING: IEEEtran.bst: No hyphenation pattern has been}%
\typeout{** loaded for the language `#1'. Using the pattern for}%
\typeout{** the default language instead.}%
\else
\language=\csname l@#1\endcsname
\fi
#2}}
\providecommand{\BIBdecl}{\relax}
\BIBdecl

\bibitem{lavie1997janus}
A.~Lavie, A.~Waibel, L.~Levin, M.~Finke, D.~Gates, M.~Gavalda, T.~Zeppenfeld, and P.~Zhan, ``Janus-iii: Speech-to-speech translation in multiple languages,'' in \emph{1997 IEEE International Conference on Acoustics, Speech, and Signal Processing}, vol.~1.\hskip 1em plus 0.5em minus 0.4em\relax IEEE, 1997, pp. 99--102.

\bibitem{nakamura2006atr}
S.~Nakamura, K.~Markov, H.~Nakaiwa, G.-i. Kikui, H.~Kawai, T.~Jitsuhiro, J.-S. Zhang, H.~Yamamoto, E.~Sumita, and S.~Yamamoto, ``The atr multilingual speech-to-speech translation system,'' \emph{IEEE Transactions on Audio, Speech, and Language Processing}, vol.~14, no.~2, pp. 365--376, 2006.

\bibitem{huang2023holistic}
W.-C. Huang, B.~Peloquin, J.~Kao, C.~Wang, H.~Gong, E.~Salesky, Y.~Adi, A.~Lee, and P.-J. Chen, ``A holistic cascade system, benchmark, and human evaluation protocol for expressive speech-to-speech translation,'' in \emph{ICASSP 2023-2023 IEEE International Conference on Acoustics, Speech and Signal Processing (ICASSP)}.\hskip 1em plus 0.5em minus 0.4em\relax IEEE, 2023, pp. 1--5.

\bibitem{lee-etal-2022-direct}
\BIBentryALTinterwordspacing
A.~Lee, P.-J. Chen, C.~Wang, J.~Gu, S.~Popuri, X.~Ma, A.~Polyak, Y.~Adi, Q.~He, Y.~Tang, J.~Pino, and W.-N. Hsu, ``Direct speech-to-speech translation with discrete units,'' in \emph{Proceedings of the 60th Annual Meeting of the Association for Computational Linguistics (Volume 1: Long Papers)}, S.~Muresan, P.~Nakov, and A.~Villavicencio, Eds.\hskip 1em plus 0.5em minus 0.4em\relax Dublin, Ireland: Association for Computational Linguistics, May 2022, pp. 3327--3339. [Online]. Available: \url{https://aclanthology.org/2022.acl-long.235}
\BIBentrySTDinterwordspacing

\bibitem{popuri2022enhanced}
S.~Popuri, P.-J. Chen, C.~Wang, J.~Pino, Y.~Adi, J.~Gu, W.-N. Hsu, and A.~Lee, ``Enhanced direct speech-to-speech translation using self-supervised pre-training and data augmentation,'' 2022.

\bibitem{jia2022translatotron}
Y.~Jia, M.~T. Ramanovich, T.~Remez, and R.~Pomerantz, ``Translatotron 2: High-quality direct speech-to-speech translation with voice preservation,'' in \emph{International Conference on Machine Learning}.\hskip 1em plus 0.5em minus 0.4em\relax PMLR, 2022, pp. 10\,120--10\,134.

\bibitem{jia2019direct}
Y.~Jia, R.~J. Weiss, F.~Biadsy, W.~Macherey, M.~Johnson, Z.~Chen, and Y.~Wu, ``Direct speech-to-speech translation with a sequence-to-sequence model,'' \emph{arXiv preprint arXiv:1904.06037}, 2019.

\bibitem{lee2021textless}
A.~Lee, H.~Gong, P.-A. Duquenne, H.~Schwenk, P.-J. Chen, C.~Wang, S.~Popuri, Y.~Adi, J.~Pino, J.~Gu \emph{et~al.}, ``Textless speech-to-speech translation on real data,'' \emph{arXiv preprint arXiv:2112.08352}, 2021.

\bibitem{lee2022direct}
A.~Lee, P.-J. Chen, C.~Wang, J.~Gu, S.~Popuri, X.~Ma, A.~Polyak, Y.~Adi, Q.~He, Y.~Tang, J.~Pino, and W.-N. Hsu, ``Direct speech-to-speech translation with discrete units,'' 2022.

\bibitem{huang2023transpeech}
R.~Huang, J.~Liu, H.~Liu, Y.~Ren, L.~Zhang, J.~He, and Z.~Zhao, ``Transpeech: Speech-to-speech translation with bilateral perturbation,'' 2023.

\bibitem{barrault2023seamless}
L.~Barrault, Y.-A. Chung, M.~C. Meglioli, D.~Dale, N.~Dong, M.~Duppenthaler, P.-A. Duquenne, B.~Ellis, H.~Elsahar, J.~Haaheim \emph{et~al.}, ``Seamless: Multilingual expressive and streaming speech translation,'' \emph{arXiv preprint arXiv:2312.05187}, 2023.

\bibitem{do2016preserving}
Q.~T. Do, T.~Toda, G.~Neubig, S.~Sakti, and S.~Nakamura, ``Preserving word-level emphasis in speech-to-speech translation,'' \emph{IEEE/ACM Transactions on Audio, Speech, and Language Processing}, vol.~25, no.~3, pp. 544--556, 2016.

\bibitem{do2018sequence}
Q.~T. Do, S.~Sakti, and S.~Nakamura, ``Sequence-to-sequence models for emphasis speech translation,'' \emph{IEEE/ACM Transactions on Audio, Speech, and Language Processing}, vol.~26, no.~10, pp. 1873--1883, 2018.

\bibitem{waibel2023face}
A.~Waibel, M.~Behr, D.~Yaman, F.~I. Eyiokur, T.-N. Nguyen, C.~Mullov, M.~A. Demirtas, A.~Kantarci, S.~Constantin, and H.~K. Ekenel, ``Face-dubbing++: Lip-synchronous, voice preserving translation of videos,'' in \emph{2023 IEEE International Conference on Acoustics, Speech, and Signal Processing Workshops (ICASSPW)}.\hskip 1em plus 0.5em minus 0.4em\relax IEEE, 2023, pp. 1--5.

\bibitem{diwan2023unit}
A.~Diwan, A.~Srinivasan, D.~Harwath, and E.~Choi, ``Unit-based speech-to-speech translation without parallel data,'' \emph{arXiv preprint arXiv:2305.15405}, 2023.

\bibitem{nachmani2023translatotron}
E.~Nachmani, A.~Levkovitch, Y.~Ding, C.~Asawaroengchai, H.~Zen, and M.~T. Ramanovich, ``Translatotron 3: Speech to speech translation with monolingual data,'' \emph{arXiv preprint arXiv:2305.17547}, 2023.

\bibitem{desplanques2020ecapa}
B.~Desplanques, J.~Thienpondt, and K.~Demuynck, ``Ecapa-tdnn: Emphasized channel attention, propagation and aggregation in tdnn based speaker verification,'' \emph{arXiv preprint arXiv:2005.07143}, 2020.

\bibitem{hsu2021HuBERT}
W.-N. Hsu, B.~Bolte, Y.-H.~H. Tsai, K.~Lakhotia, R.~Salakhutdinov, and A.~Mohamed, ``Hubert: Self-supervised speech representation learning by masked prediction of hidden units,'' 2021.

\bibitem{davis1980comparison}
S.~Davis and P.~Mermelstein, ``Comparison of parametric representations for monosyllabic word recognition in continuously spoken sentences,'' \emph{IEEE transactions on acoustics, speech, and signal processing}, vol.~28, no.~4, pp. 357--366, 1980.

\bibitem{wang2021voxpopuli}
C.~Wang, M.~Rivière, A.~Lee, A.~Wu, C.~Talnikar, D.~Haziza, M.~Williamson, J.~Pino, and E.~Dupoux, ``Voxpopuli: A large-scale multilingual speech corpus for representation learning, semi-supervised learning and interpretation,'' 2021.

\bibitem{kong2020hifigan}
J.~Kong, J.~Kim, and J.~Bae, ``Hifi-gan: Generative adversarial networks for efficient and high fidelity speech synthesis,'' 2020.

\bibitem{wang2018style}
Y.~Wang, D.~Stanton, Y.~Zhang, R.~Skerry-Ryan, E.~Battenberg, J.~Shor, Y.~Xiao, F.~Ren, Y.~Jia, and R.~A. Saurous, ``Style tokens: Unsupervised style modeling, control and transfer in end-to-end speech synthesis,'' 2018.

\bibitem{skerryryan2018endtoend}
R.~Skerry-Ryan, E.~Battenberg, Y.~Xiao, Y.~Wang, D.~Stanton, J.~Shor, R.~J. Weiss, R.~Clark, and R.~A. Saurous, ``Towards end-to-end prosody transfer for expressive speech synthesis with tacotron,'' 2018.

\bibitem{ren2022fastspeech}
Y.~Ren, C.~Hu, X.~Tan, T.~Qin, S.~Zhao, Z.~Zhao, and T.-Y. Liu, ``Fastspeech 2: Fast and high-quality end-to-end text to speech,'' 2022.

\bibitem{sun2020fullyhierarchical}
G.~Sun, Y.~Zhang, R.~J. Weiss, Y.~Cao, H.~Zen, and Y.~Wu, ``Fully-hierarchical fine-grained prosody modeling for interpretable speech synthesis,'' 2020.

\bibitem{Kahn_2020}
\BIBentryALTinterwordspacing
J.~Kahn, M.~Riviere, W.~Zheng, E.~Kharitonov, Q.~Xu, P.~Mazare, J.~Karadayi, V.~Liptchinsky, R.~Collobert, C.~Fuegen, T.~Likhomanenko, G.~Synnaeve, A.~Joulin, A.~Mohamed, and E.~Dupoux, ``Libri-light: A benchmark for asr with limited or no supervision,'' in \emph{ICASSP 2020 - 2020 IEEE International Conference on Acoustics, Speech and Signal Processing (ICASSP)}.\hskip 1em plus 0.5em minus 0.4em\relax IEEE, May 2020. [Online]. Available: \url{http://dx.doi.org/10.1109/ICASSP40776.2020.9052942}
\BIBentrySTDinterwordspacing

\bibitem{veaux2017english}
C.~Veaux, J.~Yamagishi, K.~MacDonald \emph{et~al.}, ``English multi-speaker corpus for cstr voice cloning toolkit,'' \emph{The Centre for Speech Technology Research (CSTR), University of Edinburgh}, 2017.

\bibitem{ljspeech17}
K.~Ito and L.~Johnson, ``The lj speech dataset,'' \url{https://keithito.com/LJ-Speech-Dataset/}, 2017.

\bibitem{ott2019fairseq}
M.~Ott, S.~Edunov, A.~Baevski, A.~Fan, S.~Gross, N.~Ng, D.~Grangier, and M.~Auli, ``fairseq: A fast, extensible toolkit for sequence modeling,'' \emph{arXiv preprint arXiv:1904.01038}, 2019.

\bibitem{wang2020fairseq}
C.~Wang, Y.~Tang, X.~Ma, A.~Wu, S.~Popuri, D.~Okhonko, and J.~Pino, ``Fairseq s2t: Fast speech-to-text modeling with fairseq,'' \emph{arXiv preprint arXiv:2010.05171}, 2020.

\bibitem{wang2021fairseq}
C.~Wang, W.-N. Hsu, Y.~Adi, A.~Polyak, A.~Lee, P.-J. Chen, J.~Gu, and J.~Pino, ``fairseq s\^{} 2: A scalable and integrable speech synthesis toolkit,'' \emph{arXiv preprint arXiv:2109.06912}, 2021.

\bibitem{polyak2021speech}
A.~Polyak, Y.~Adi, J.~Copet, E.~Kharitonov, K.~Lakhotia, W.-N. Hsu, A.~Mohamed, and E.~Dupoux, ``Speech resynthesis from discrete disentangled self-supervised representations,'' \emph{arXiv preprint arXiv:2104.00355}, 2021.

\end{thebibliography}

\end{document}